# Efficient Global Optimization of Non-differentiable, Symmetric Objectives for Multi Camera Placement

Maria L. Hänel, Carola B. Schönlieb

*Abstract*—We propose a novel iterative method for optimally placing and orienting multiple cameras in a 3D scene. Sample applications include improving the accuracy of 3D reconstruction, maximizing the covered area for surveillance, or improving the coverage in multi-viewpoint pedestrian tracking. Our algorithm is based on a block-coordinate ascent combined with a surrogate function and an exclusion area technique. This allows to flexibly handle difficult objective functions that are often expensive and quantized or non-differentiable. The solver is globally convergent and easily parallelizable. We show how to accelerate the optimization by exploiting special properties of the objective function, such as symmetry. Additionally, we discuss the trade-off between non-optimal stationary points and the cost reduction when optimizing the viewpoints consecutively.

*Index Terms*—NETW networked sensor fusion and decisions; OPTO imaging sensors; APPL environmental monitoring and control, robotics and automation; OTHER global optimum, multiple camera placement, non-differentiable objective, symmetric objective;

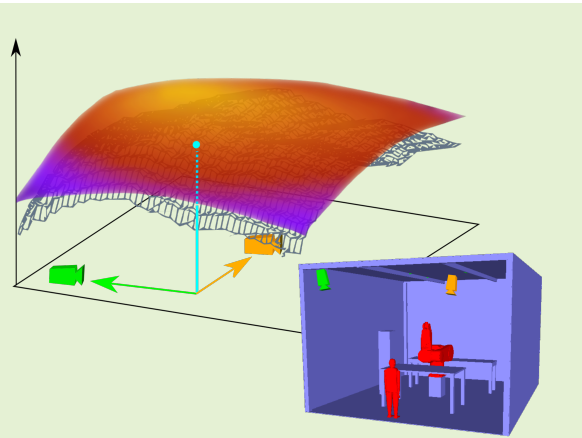

Fig. 1. Placing and orienting multiple cameras at the global optimum of a given CAD environment

## I. Introduction

CAMERA NETWORK applications range from validating robot swarms [1] over object tracking [2] to 3D reconstruction [3]. They are often applied for critical tasks, such as ensuring the integrity of a human being in industrial robotics [4] or in suveillance systems [5] or supporting a decision with considerable financial consequences at professional sport events [6]. Thus, the deployment of cameras often comes with the responsibility to build a failure-resistant system. This system will be more precise and the relevant regions of the environment are better covered, the better the cameras are placed and oriented. The purpose of this work is to find globally optimal positions and orientations of the cameras in a given 3D environment.

Optimal camera placement can be phrased as an optimization problem. The parameters of the cameras which need to be adjusted, such as positions and orientations, are the *variables* of the optimization. These are constrained by the mounting area of the environment and possible mounting directions. The *domain* $\mathcal{D}$ is the variable space for all the cameras of the network. A *solver* iterates over the domain to find the best camera constellation. In each iteration step of the solver, the quality of the current *camera constellation* (i.e. positions and orientations of multiple cameras) needs to be measured. In general, the quality of a camera constellation is encoded in the *objective function* $f : \mathcal{D} \to \mathbb{R}$, a real-valued function such as the volume of the camera's united field of view (FOV) or a reconstruction error. In this work we propose an iterative, numerical algorithm (Fig. 1) that efficiently produces a set of optimal camera positions and orientations, i.e. a solver for a non-linear, real-world problem optimizing $f$.

### A. Properties of the Problem

*1) Positive:* The objective function $f$ that measures the quality of a camera constellation has some properties that help to accelerate the optimization:

Submission date: January 15, 2021

Dr. Hänel acknowledges the support from the German Academic Exchange Society DAAD.

CBS acknowledges support from the Leverhulme Trust project on 'Breaking the non-convexity barrier', the Philip Leverhulme Prize, the Royal Society Wolfson Fellowship, the EPSRC grants EP/S026045/1 and EP/T003553/1, EP/N014588/1, EP/T017961/1, the Wellcome Innovator Award RG98755, European Union Horizon 2020 research and innovation programmes under the Marie Skodowska-Curie grant agreement No. 777826 NoMADS and No. 691070 CHiPS, the Cantab Capital Institute for the Mathematics of Information and the Alan Turing Institute.

Cambridge Image Analysis, Department of Applied Mathematics and Theoretical Physics, University of Cambridge, Wilberforce Road, Cambridge CB3 0WA, UK (e-mail: maria.l.haenel@gmail.com, cbs31@cam.ac.uk)

*a) Symmetry:* Most of the time, cameras of the same type are installed, so the objective function is invariant under the permutation of cameras. $f$ is said to be *symmetrical*.

*b) Prior:* The field of view of a camera behaves just as in human vision, so relatively good, non-occluded camera positions may be found to initialize the optimization.

*c) Subspaces:* Once the quality has been evaluated for the first camera constellation, readjusting cameras consecutively (greedy method) reduces the costs of the subsequent evaluations since the field of view of the rest of the cameras will not change. In terms of optimization, the parameters of a single camera of the network lie in a *subspace* of the domain.

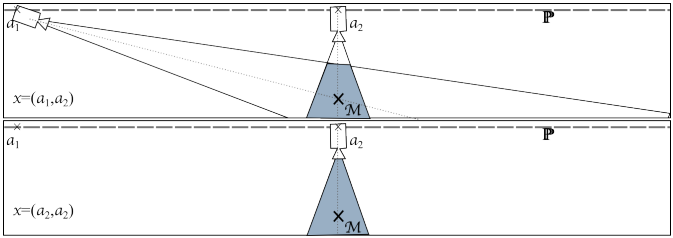

Fig. 3. Example of non-optimal stationary point: Top: The initial adjustment for camera 1 is $a_1 \in \mathbb{P}$ and for camera 2 is $a_2 \in \mathbb{P}$. Both cameras are placed in the domain $\mathcal{D} \in \mathbb{P}^2$ (on dashed line always directed to $M$). Our goal is to maximize the volume of the intersection of the camera's FOV (grey). The global maximum is adopted when both cameras are all the way to the left or right $\pm(a_1, a_1)$.
Bottom – greedy method: First optimize camera 1 (move the left camera to $a_2$), then optimize camera 2. After the first step, both cameras are in $a_2$, since this is the optimum on the subspace. Even after changing the subspace and optimizing camera 2, the optimal solution is again $(a_2, a_2)$ (since the function is symmetric). However, moving both cameras at the same time left or right $((a_2, a_2) + \epsilon \cdot (1, 1)$ with any $\epsilon > 0$ or $< 0$) will increase the volume of the FOV. Thus, the stationary point $(a_2, a_2)$ is neither a local nor a global maximum.

*2) Negative:* The objective $f$ has some particular properties that are challenging for most of the existing solvers:

*a) Stationary solutions:* Alternately placing single cameras (greedy method) has been adopted tolerating the fact that it also leads to *stationary solutions* which are neither local nor global optima (Fig. 3).

*b) Expensive:* When measuring the reconstruction error or the field of view of several cameras in a CAD environment, $f$ comprises geometrical operations and is expensive.

*c) Black-box:* The geometrical simulation renders $f$ to be a black box, i.e. operations other than function evaluations are hardly possible. Thus, $f$ cannot be decomposed in its additively seperable parts or derived analytically. Approximations such as the difference quotient lead to increased costs.

*d) Non-smooth:* In [7], $f$ is piecewise non-differentiable and non-linear, a gradient does not always exist.

*e) Non-convex:* It is also non-convex with multiple local optima [7]. Searching simply in direction of the gradient will only yield a local but not global optimum.

*f) Quantized:* In [8], $f$ is even discontinuous and in [9] the discretized voxel set leads to a quantized $f$. This means that even if the gradient existed, it might be zero almost everywhere.

## B. Contributions and Content

Due to these complications, the problem is usually simplified by the reduction of the domain's dimenisions or constraints, e.g. by the choice of camera positions out of predefined sets, by arranging the mounting spots around the interesting objects, by choosing a two-dimensional surveillance space, or accepting statinary solutions. Instead of simplifying the objective to overcome these difficulties, a suitable solver evaluates the objective function along subspaces of the domain, smoothens the objective $f$, recognizes the symmetric counterparts, caches the function values $f(x)$ as to not evaluate them again, and finds a *globally* optimal solution in the end. We achieve this by combining the following methods:

*a) Subspace decomposition/BCA:* The domain of the problem is decomposed into the subspaces, each including only the parameters of a single camera. This is called *domain decomposition*. The subset of variables of a single camera is called a *block* of variables. We will use a method in which each block is optimized in turns and is called *block coordinate ascent/descent* (BCA) and we will discuss suboptimal stationary points.

*b) Surrogate/RBF:* A non-smooth, quantized or black-box function still needs to be smoothened or approximated such that it delivers a gradient. In optimization, a *surrogate* of the objective function is a differentiable function which interpolates previous evaluation points and whose function value and derivative are cheaply evaluated. As soon as a high quality iterate according to the surrogate is found, the real objective is evaluated and the surrogate is updated with the new evaluation point. This way we cache function values of previos evaluations and establish a gradient.

*c) Exclusion Areas/EAM:* With a BCA the optimization may yield suboptimal stationary solutions. In this work, we adopt a strategy to exclude areas with already evaluated sample pairs in order to force to avoid stationary points.

This solver can be computed in parallel and is globally convergent on a continuous domain for various objectives. We will demonstrate the originality of our method and a general demand for an efficient solver for optimal camera placement in Section II. Section III elaborates on the design of the solver. Elaborate synthetic and realistic experiments in Section IV show that the solver is faster than global solvers and more accurate than local solvers by covering the complete domain.

## II. RELATED WORK

We will first discuss previous camera placement approaches and their objective functions in Section II-A. Then, an overview on useful solvers for these problems follows in Section II-B.

## A. Optimal Camera Placement

The optimal camera constellation for one application is not necessarily the optimal solution in a different field, which is why research has been conducted for the following objective functions:

3*1) General problems:* In case the application is unknown, a general form of objective function needs to be used. The *Art Gallery Problem* is concerned with minimizing the number of cameras such that a set of objects or an area of the environment is observed completely [10]–[19]. The reverse problem, the Maximum Set Covering Location Problem maximizes the area or the number of objects/paths observed by a given number of cameras, [8], [20]–[23].

*2) Application specific problems:* In case the camera network needs to satisfy a specific quality function, a general objective function would miss some particulars. There are online methods that dynamically adjust the cameras while performing a (robotic) task, such as eye-in-hand movement [24]–[26], scene modelling [27], object reconstruction [28], active stereo tracking [29], decentralized tracking [30], visual servoing [31], or multi-robot formation [32]. In contrast to these, offline methods are designed for optimizing and attaching the cameras before executing the task without [33]–[35] and with regarding obstacles [7], [9], [15], [16], [36], [37] or risk maps [38] or uncertainty [8] in the environment.

*3) Typical optimal placement:* The represented authors give an extended problem or visibility analysis and experiments for optimal camera placement and use popular optimization techniques including greedy methods [8], [19], [39], [40] or hillclimbing [8] or stochastic methods [7], [9], [41]. For global optimality, some authors formulate these problems as a binary program with different types of visibility and connectivity constraints [42]–[45] and solve it with Branch and Bound [11], [46]. Nonlinear constraints such as the visibility of a camera are typically precomputed and they choose camera positions out of a predefined set. In order to compute the objective online in this work and use a continuous domain, we combine non-linear optimization methods to increase the efficiency of the optimization of a symmetrical, expensive, black-box objective which is easier evaluated on camera specific subspaces. Some authors have optimized a three-dimensional problem on sub-spaces but they either provide gradient information [47] or do not discuss non-optimal, stationary points [8], [26], [48], [49].

### B. Optimization

In order to compute a global optimum, two main strategies generally exist, iterating through each local optimum II-B.1 or, in case the local optima cannot be found deterministically, randomly sampling the domain II-B.2. Both II-B.1 and II-B.2 present weaknesses for optimal camera placement. Therefore, we suggest the use of a BCA II-B.3 in combination with a surrogate II-B.4 instead in a combination II-B.5:

*1) Deterministic methods:* Some methods such as the Nelder-Mead-Simplex [50], [51], Interior Point Filter Line Search [52], Method of Moving Assymptotes [53], or the Sequential Quadratic Program [54], [55] use the objective function's convexity, separability, or derivative information to prove the convergence to a local optimum. Properties which may not be adopted by the objective function or information which may not be known, compare Section I-A.

*2) Randomized methods:* Randomized solvers use the fact that a global optimum can be found by sampling an arbitrary continuous function dense in a compact domain [56]. Two examples of dense sampling in randomized methods are the Ant Colony Algorithm [57] and Simulated Annealing [58], which have been used in global optimal camera placement in [9] and [7], respectively. However, for dense iterates a huge number of iterations is needed and the cost of the objective is multiplied by this number. For instance, in [7] cameras are placed in a 2D environment and for "very high dimensional spaces ($> 8$ <parameters>), although the algorithm provided reasonably good solutions very quickly, it sometimes took several hours to jump to a better solution".

*3) Subspace decomposition/BCA:* The convergence of a coordinate search, block coordinate ascent/descent (BCA), block-nonlinear Gauss Seidel, alternating minimization, and domain decomposition – as the optimization with a subspace decomposition is called often – has been studied for both overlapping [59], [60] and non-overlapping subspaces [61] and under various assumptions, e.g., for strictly convex, quadratic or separable functions [62]. The problem that the subspaces may not lie in the gradient direction of the objective [63] has been addressed by rules that choose the order and direction of the subspaces [64], [65] the latter is parallelized for large data [66]. In optimal camera placement, the subspaces need to be parallel to the coordinate axes of the blocks, orthogonal to each other, and non-overlapping (according to the variables of each camera). Thus, a rule for the choice of the subspace direction cannot be applied. Moreover, there are points at which the choice of the subspace is irrelevant due to the symmetry of the objective function (Fig. 3).

*4) Surrogate/RBF:* For non-linear optimization, the surrogate needs to interpolate *scattered* evaluation points, i.e. points which are not arranged on a grid, compare [67], [68] for multivariate interpolation and approximation methods. Some methods approximate a subset of evaluation points in a piecewise manner (subsets may be achieved by Delaunay triangulation [69]). Alternatively, a radial basis function interpolant (RBF) [70]–[72] interpolates the whole set of evaluation points without triangulation (Fig. 4). The authors of [73], [74], have developed optimization methods with RBFs as surrogates. Powell [71] developed a method to add evaluation points subsequently to the RBF, like Newton's subsequent interpolation method.

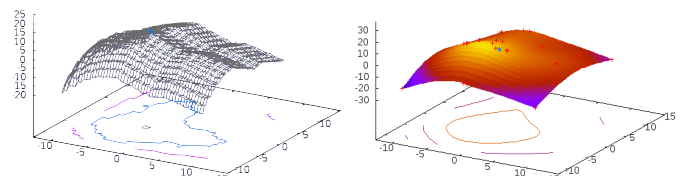

Fig. 4. Illustration of a quantized objective function (left) and the corresponding radial basis function (RBF, right) after an interpolation using 21 scattered evaluation points (red crosses).

*5) Combination:* The combination of BCAs and RBFs has been studied in [75], [76] for large problems under the assumption that not all variables are equally relevant. The authors' global search method dynamically adjusts the size of a block whose variables are perturbed in each iteration.

However, in optimal camera placement, each block/subspace (corresponding to the variables of one camera) is equally relevant and usually contains a similar number of variables.

### C. Problem Definition

Let $\mathbb{P}$ denote the variable space of a single camera, for instance one can use $\mathbb{P} = \{(p,o) \mid p \in \mathbb{R}^3, o \in [0, 2\pi] \times [0, \pi]\}$ consisting of the position $p$ and the yaw and pitch of the camera $o$. For $M \in \mathbb{N}$ cameras $a_1 \in \mathbb{P}_1, ..., a_M \in \mathbb{P}_M$, we call $x := (a_1, ..., a_M)$ a *camera constellation* or the *variable vector* and the space $\mathcal{D} := \mathbb{P}_1 \times ... \times \mathbb{P}_M$ the *domain*.
The problem of finding the optimal camera constellation in the domain with a fixed number of cameras is denoted by

$$\text{Find } \underset{x \in \mathcal{D}}{\arg\max} f(x). \tag{1}$$

In this work, the objective function $f$ is decomposed by (2) into $M$ camera specific functions $\sigma_m$ which merely depend on the parameters of a single camera $a_m \in \mathbb{P}_m$. The *observation set* $\mathbb{A}_m$, $m \in \{1, \dots, M\}$ resembles the set of all objects, features, voxels, etc. which can be captured by the $m$-th camera and is kept abstract intentionally.

$$f(x) = g \circ (\sigma_1, ..., \sigma_M)^T (x) \tag{2}$$
$$g: \quad (\mathbb{A}_1 \times ... \times \mathbb{A}_M) \to \mathbb{R} \tag{3}$$
$$\sigma_m: \quad \mathbb{P}_m \to \mathbb{A}_m \text{ for all } m \in \{1, \dots, M\}. \tag{4}$$

In practice, the decomposition of $f$ in camera specific functions $\sigma_m$ and the use of a *fusion function* $g$ is a valid assumption which can be applied in many optimal camera placement scenarios. In Table I, the decomposition of sample problems from [7]–[9] are given. In these examples, all the cameras are allowed to be placed at the same mounting spots $\mathbb{P} := \mathbb{P}_1 = \dots = \mathbb{P}_M$. The same sensor model is used for each camera $\sigma := \sigma_1 = \dots = \sigma_M$ which can make the same observations $\mathbb{A} := \mathbb{A}_1 = \dots = \mathbb{A}_M$. Thus, their objective functions are symmetric in all the subspaces.

TABLE I
EXAMPLES FOR OPTIMAL CAMERA PLACEMENT DECOMPOSITIONS

|     | A single cam.'s observation $\sigma_m$ | Represent. $\mathbb{A}_m$ | Fusion $g$ |
|-----|----------------------------------------|---------------------------|------------|
| [9] | Simulation of FOV in the CAD model | Voxel set | Set operations of $\sigma_m$ and distance to robot model |
| [7] | Determination of the area of multiple objects' region of occlusion | Real value | Probability of an object being visible from all sensors |
| [8] | Analysis of observation of pedestrian paths | Vector of path observabilities | Sum of vectors including a clause directing new cams. away from already observed paths |

## III. PROPOSED SOLVER

Our solver is customised for problems (1) in the general form (2) and (4) that fulfill the properties outlined in Sections I-A and II-C. We combine a block coordinate ascent BCA (Section III-B) and an exclusion area method EAM (Section III-C) with a surrogate (Section III-A) of the objective function as proposed in Section I-B. Although a natural choice, this has not been considered for optimal camera placement before, cf. Section II.

### A. Surrogate

In optimization, a *response surface model* or *surrogate* of an objective function $f: \mathcal{D} \to \mathbb{R}$ with domain $\mathcal{D} \subset \mathbb{R}^n$, $n \in \mathbb{N}$ is an inexpensive function $\bar{f}: \mathcal{D} \to \mathbb{R}$ that interpolates (possibly) scattered evaluation points $(s_1, f_1), \dots, (s_K, f_K) \in \mathcal{D} \times \mathbb{R}$ for which an objective value is already known $f_k := f(s_k)$, $k \in \{1, \dots, K\}$. In this publication, a radial basis function interpolant [70], [71] as illustrated in Fig. 4 is used as a surrogate as in [77].

Let $||.||$ denote the Euclidean norm. Let $\phi: \mathbb{R}_o^+ \to \mathbb{R}$ be a continuously differentiable function with $\phi(0) = 0$. Let $\Pi_m^n$ be the linear space of polynomials of degree less than or equal to $m \in \mathbb{N}$ with $n$ variables. A real function $\bar{f}: \mathcal{D} \to \mathbb{R}$ with

$$\bar{f}(x) := \sum_{k=1}^{K} \omega_k \phi(||x - s_k||) + p(x), \quad x \in \mathcal{D} \tag{5}$$

is called a *radial basis function interpolant* (RBF) of $s_1, \dots, s_K$ on $f$ if weights $\omega_1, \dots, \omega_K \in \mathbb{R}$ and a polynomial $p \in \Pi_m^n$ exist with

$$f_k = \bar{f}(s_k) \text{ for all } k \in \{1, \dots, K\} \tag{6}$$

$$\mathbf{0} = \sum_{k=1}^{K} \omega_k q(s_k) \quad \forall q \in \Pi_m^n. \tag{7}$$

The function $\phi$ is called the *kernel* of the RBF. In this work, we use a linear polynomial with a thin plate spline as a kernel:

$$\phi(r) := r^2 \log(r), \ p(x) := \nu \cdot x + \nu_0, \quad r, \nu \in \mathbb{R}^n, \nu_0 \in \mathbb{R} \tag{8}$$

With this kernel, the interpolant $\bar{f}(x)$ (5) is continuously differentiable. Notice that the argument of the kernel is the radially symmetric distance of $x$ to a candidate $s_k$. The weights $\omega, \nu,$ and $\nu_0$ in (5) are the variables of the system of linear equations (6), which is underdetermined with $K + n + 1$ variables and $n$ constraints. To enforce uniqueness of the interpolant and regularity of the system's matrix, the conditions (7) are introduced.

In our solver the surrogate is optimized instead of the expensive objective function. When a good solution for the surrogate $s_K$ is found, the actual (costly) objective function $f(s_K)$ is evaluated and the surrogate is updated by the new evaluation point. The update of the RBF is depicted in Algorithm 1.

---

**Algorithm 1** $(\bar{f}, x_o, f_o) \leftarrow Update(f, s_1, \dots, s_{K-1}, s_K)$
Outputs: Update of surrogate $\bar{f}$ with each eval. point $(s_k, f(s_k))$, $k \in \{1, \dots, K\}$, update of maximum pair $(x_o, f_o)$.

1: $\mathbf{f_K} \leftarrow \mathbf{f(s_K)}$                 //expensive function evaluation

2: Solve eq. syst. (6) and (7) with variables $\omega, \nu, \nu_0$

3: $f_o \leftarrow \max(f_1, \dots, f_K)$
4: $x_o \leftarrow \arg\max(f_1, \dots, f_K)$
5: **return** $(\bar{f}(\omega, \nu, \nu_0), x_o, f_o)$

---

The objective function $f$ is only evaluated in Algorithm 1 (Line 1). This way, all previous evaluation points are cached in the surrogate and the function value as well as the gradient of intermediate variable vectors $x$ ($\neq s_k, \forall k = 1, \ldots, K$) can be approximated easily. Moreover, symmetric points of already evaluated variable vectors can be updated with the same objective value, in Line 2 by adding an additional row and column to the equation system.

## B. Block Coordinate Ascent

For finding the next evaluation point (a.k.a. next good camera constellation) with unknown objective value to be integrated into the surrogate we employ the following approach. Within a computational optimization procedure, the $n_m \in \mathbb{N}$ variables of a single camera $a_m \in \mathbb{P}_m, m \in \{1, \ldots, M\}$ represent a *block* of the whole variable vector $x$. The procedure of optimizing blocks of variables in turns is called *Block Coordinate Ascent/Descent* (BCA). The decomposed domain $\mathcal{D} := \mathbb{P}_1 \times \ldots \times \mathbb{P}_M$ has the dimension $n = n_1 + \ldots + n_M$. We decompose the identity matrix $\mathbf{1}_n$ into the partitions $U_m$:

$$\mathbf{1}_n := (U_1, \ldots, U_M) \in \mathbb{R}^{n \times n}, U_m \in \mathbb{R}^{n \times n_m}, m = 1, \ldots, M.$$

Instead of solving the problem (1) on the whole domain $\mathcal{D}$, a sequence of smaller problems is solved iteratively. Choose the initial variable vector $x^{(0)} := U_1 a_1^{(0)} + \ldots + U_M a_M^{(0)} \in \mathbb{P}_1 \times \ldots \times \mathbb{P}_M$, e.g. $x^{(0)} = \mathbf{0}$ and subsequently solve ($i \to i+1$):

$$u_1 := \underset{v_1 \in \mathbb{P}_1}{\operatorname{argmax}} f(U_1 v_1 \qquad\qquad\qquad +x^{(i)}) \quad (9)$$

$$u_2 := \underset{v_2 \in \mathbb{P}_2}{\operatorname{argmax}} f(U_2 v_2 \quad + U_1 u_1 \qquad +x^{(i)}) \quad (10)$$

$$\vdots$$

$$u_M := \underset{v_M \in \mathbb{P}_M}{\operatorname{argmax}} f(U_M v_M + \sum_{m=1}^{M-1} U_m u_m + x^{(i)}) \quad (11)$$

$$x^{(i+1)} := U_1 u_1 + \ldots + U_M u_M + x^{(i)}, \text{ and go to (9)} \quad (12)$$

The only variable parameters in each line $m = 1 \cdots M$ are the variables of the $m$-th block $v_m \in \mathbb{P}_m$, the parameters of the remaining cameras are constant. We call the optimization in each line a *subspace optimization*. Since the previous evaluation point is $x^{(i)} = (a_1, \ldots, a_M)$, the optimal solution in this subspace is

$$(a_1 + u_1, \ldots, a_{m-1} + u_{m-1}, \ a_m + u_m, \ a_{m+1}, \ldots, a_M).$$

The result $u_m \in \mathbb{P}_m$ resembles an *offset* between $x^{(i)}$ and the optimal solution in the subspace $\mathbb{P}_m$. In the last step (12), the communication between the subspaces is established by adding the offsets of the recent iteration step to get a new variable vector $x^{(i+1)}$.

By omitting the term $\sum_{m=1} U_m u_m$ in each line (9)–(11), each subspace optimization has the same starting point, namely $x^{(i)}$. Thereby, each subspace optimization does not depend on the results of the subspaces before and thus the lines (9)–(11) can be computed in parallel.

*1) Complexity of one iteration step:* For a problem of the general form (1)-(4) with the objective function $f(x) = g \circ (\sigma_1, \ldots, \sigma_M)^T(x)$, let the cost for the determination of the camera specific function $\sigma_m$ be denoted by $c_\sigma$ and let $c_g$ denote the cost for the fusion $g$. Let $I_m \in \mathbb{N}$ be the number of times that the function $f(x)$ is evaluated in the optimization of subspace $\mathbb{P}_m$. The evaluation $f(x^{(i)})$ with the complete set of $M$ camera observations $\sigma_1(a_1), \ldots, \sigma_M(a_M)$ needs to be calculated only in the first function evaluation with $\mathcal{O}(M \cdot c_\sigma + c_g)$. All except $\sigma(a_m)$ are constant in the rest of the $I_m - 1$ subspace evaluations, resulting in $\mathcal{O}((I_m - 1) \cdot (c_\sigma + c_g))$. With the maximum number of subspace optimization steps $I_0 = \max_{m \in \{1, \ldots, M\}} I_m$, the complexity of the full iteration step ($i \to i+1$) would already be reduced to $\mathcal{O}(M(Mc_\sigma + c_g + (I_0 - 1)c_\sigma + c_g)))$. In case of a parallel computation, $f(x^{(i)})$ is only called once in the whole iteration step. Let $T \in \mathbb{N}$ be the number of possible parallel threads, reducing the complexity further:

$$\mathcal{O}\left(\left\lceil \frac{M}{T} \right\rceil c_\sigma + c_g + \left\lceil \frac{M}{T} \right\rceil (I_0 - 1)(c_\sigma + c_g)\right). \quad (13)$$

*2) Stationary points:* Let $x_* \in \mathcal{D}$ not be in the boundary of the domain. A possible termination criterion for the total BCA procedure is the arrival at a *stationary point* $x_* \in \mathcal{D}$ which holds

$$x_*^{(i)} = x_*^{(i+1)} \text{ with } u_m = 0, \ m = 1, \ldots, M \quad (14)$$

If the $m$th offset is zero $u_m = 0$ in the $m$th subspace optimization at the point $x_*$ then

1) NO direction $r = U_m v_m$ in the $m$th block $v_m \in \mathbb{P}_m$ exists with $f(r + x_*) > f(x_*)$ or
2) ($f$ differentiable) the $m$th block of the gradient $\nabla_m = (\nabla f(x_*))_m$ is zero

With these termination criteria, let us compare the stationary points and local optima depicted in Table III-B.2 in order to find suboptimal stationary points from the example in Fig. 3: In the example $x_* = (a_2, a_2)$ is a stationary point, all offsets are zero. If $f$ was differentiable in $x_*$, the gradient on all the subspaces would be zero, which renders the total gradient $\nabla f(x_*) = (\nabla_1, \ldots, \nabla_M)^T$ being zero as well. Even in case all the blocks of $x_*$ are local optima on their subspaces, $x_*$ may still be both, a local optimum or a saddle point. Is the saddle point the suboptimal stationary point we are looking for? The gradient in Fig. 3 cannot be zero, since $x_* = (a_2, a_2)$ CAN be improved in a direction $x_* + r$ with $r = \pm(1, 1)$ only it is in neither subspace $r \neq U_m v_m \in \mathbb{P}_m \ m = 1, \ldots, M$. $x_*$ is not a saddle point, it is in fact not differentiable at all.

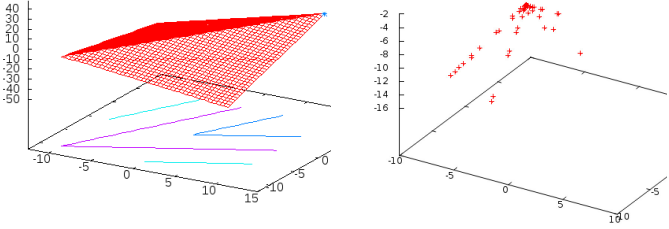

Fig. 5. Subspace optimum; Left: The maximum of the piecewise linear function (red surface) on the domain $[-10, 10] \times [-10, 10]$ is illustrated as a blue cross. Right: The intermediate iterates of the BCA on the piecewise linear function are illustrated as red crosses. The maximum of each of the two subspaces coincides but is neither a local nor global maximum nor a saddle point, since the gradient is not zero.

TABLE II
STATIONARY AND NON STATIONARY POINTS

|  | Improvement direction exists in: | | |
|---|---|---|---|
|  | $\mathbb{P}_m \subset \mathcal{D}$ | $\mathcal{D}$ not in $\mathbb{P}_m$ | does not exist |
| $\nabla f(x_*) = \mathbf{0}$ | saddle point | saddle point | local smooth optim. |
| $\nabla f(x_*) \neq \mathbf{0}$ | × | – | – |
| $\nexists \nabla f$ | × | subspace optim. | non-smooth optim. |

Stationary (grey) and non stationary points (white) can be differentiated by two criteria: Differentiability stati (rows) and the improvement direction (columns). Second line: A differentiable point with a zero gradient can only be a stationary point, either a non-optimal saddle point (if an improvement direction exists, 2nd and 3rd column) or a locally optimal smooth optimum (if no local improvement direction exists). Third line: The empty × fields are not stationary since an improvement direction exists in one of the subspaces. The empty − fields are not possible since $\nabla f \neq \mathbf{0}$ is a contradiction to no improvement direction exists (along blocks). Fourth row (non-differentiable points): In case an improvement direction exists in a direction other than the blocks, the stationary point is a non-optimal subspace optimum. If an improv. direction does not exist at all, the stationary point is an optimal non-smooth optimum.

Well known stationary points are local optima and saddle points. However, stationary points exist which are neither. A *subspace optimum* is a non-differentiable, stationary point holding (14) with an improvement direction outside the subspaces (Fig. 5).

### C. Exclusion Area Method

Computationally, local optima, saddle points and subspace optima must be considered as an optimal point for updating the surrogate since (a) no gradient exists and/or (b) no improvement direction in one of the subspaces exist that prove otherwise. Even a differentiable surrogate of the objective function in these examples may develop a saddle point or a local optimum unless a strategy is introduced that establishes global convergence instead of a convergence to a non-optimal stationary point.

**Algorithm 2** $s_{K+1} \leftarrow Search(V, \bar{f}, s_1, \ldots, s_K, \beta)$
Search for the next evaluation point depending on the search space $V$, surrogate $\bar{f}$, all previous variable vectors $s_k$, $k = 1, \ldots, K$, and the size $\beta$ of exclusion areas

1: $\Delta \leftarrow \max_{x \in \mathbb{P}} \min_{1 \leq k \leq K} \|x - s_k\|$
2: $s_{K+1} \leftarrow$ **Maximize** $\bar{f}(x)$ subj. to
3: $\quad \|x - s_k\| \geq \beta\Delta, \ k = 1, \ldots, K$
4: $\quad x \in V$
5:
6: **return** $s_{K+1}$

The basis of the global convergence of randomized solvers is a theorem by [56, Theorem 1.3] adjusted to our notation: "Let the 'optimization' region $\mathcal{D}$ be compact. Then, a global 'optimization' algorithm converges to the global 'optimum' of any continuous function iff the sequence of iterates of the algorithm is everywhere dense in $\mathcal{D}$." The randomized strategy by [77] (Algorithm 2) has been used in combination of surrogate $\bar{f}$. The vicinity of each previous variable vector defines an *exclusion area* which is excluded from further sampling in the domain in Line 3. Hereby, the size of the exclusion areas $\beta\Delta$ can be adjusted by $\beta \in [0, 1)$. The radius $\Delta \in \mathbb{R}$ in Line 1 is the largest distance between the previously chosen variable vectors $s_1, \ldots, s_K \in \mathcal{D}$ and the furthest point in the search space $V \subset \mathcal{D}$. Under these constraints, the new variable vector maximizes the surrogate $\bar{f}$ in the search space $V$.

The method of Exclusion Areas (EAM) is repeating the search and update strategies (Algorithms 2 and 1). By decreasing the size of the exclusion areas, the solver determines the next variable vector more locally in the vicinity of previous variable vectors, the choice is rather global when increasing the size. In order to use both local and global strategies, [77] has chosen to vary $\beta$ in *cycles* of $L \in \mathbb{N}$ different sizes: $< \beta_1, \ldots, \beta_L >$.

### D. RBF-BCA

A combination of the **BCA**, the EAM, and the **RBF** as a surrogate is depicted in Algorithm 3 (RBF-BCA). The solver stores the previous evaluation points $s_{1,\ldots,K}$ in a surrogate $\bar{f}$ of the objective function $f$, as to not evaluate them again and to be able to optimize on a simpler function.

The next evaluation point $s_{K+1}$ is chosen outside the exclusion areas in Line 4 and integrated into the surrogate in Line 5. The sequence from Line 6 until Line 11 constitute the BCA iteration steps (9)-(11). Beginning at the last evaluation point $s_{K+1}$, the solver searches the surrogate for the best solution in one subspace of the domain. After having found the maximum in the subspace, the subspace is changed. The expensive objective function is evaluated at only such a changing point. The Line 12 handles the BCA update (12). The evaluation point is then incorporated with its symmetrical evaluation points into the surrogate in Line 13.

The RBF-BCA is terminated if a desired density of recent variable vectors $\Delta_0 \in \mathbb{R}$ is achieved. The global convergence of the evaluation points to an optimal solution is established since $\Delta \to 0, (K \to \infty)$ with Lines 5 and 6 [78] and therefore

**Algorithm 3** RBF-BCA

BCA using a surrogate-update $Update()$ and an exclusion area strategy $Search()$ with termination criterion $\Delta_0 \in \mathbb{R}$.

1: $(\bar{f}, x_o, f_o) \leftarrow Update(f, s_1, \ldots, s_K)$
2: **while** $\Delta \leq \Delta_0$ **do**
3:     **for all** $\beta$ in $<\beta_1, \ldots, \beta_L>$ **do**
4:         $s_{K+1} \leftarrow Search(\mathcal{D}, \bar{f}, s_1, \ldots, s_K, \beta)$
5:         $(\bar{f}, x_o, f_o) \leftarrow Update(f, s_1, \ldots, s_{K+1})$
6:         **while** $s$ is not stationary point **do**
7:             **for all** subspaces $\mathbb{P}_m, m = 1, \ldots, M$ **do**
8:                 $t_m \leftarrow Search(\mathbb{P}_m, \bar{f}, s_1, \ldots, s_{K+1}, \beta)$
9:                 $(\bar{f}, x_o, f_o) \leftarrow Update(f, s_1, \ldots, s_{K+1}, t_m)$
10:             **end for**
11:         **end while**
12:         $s_{K+2} \leftarrow U_1 t_1 + \ldots U_M t_M$
13:         $(\bar{f}, x_o, f_o) \leftarrow Update(f, s_1, \ldots, s_{K+1}, s_{K+2})$
14:     **end for**
15: **end while**

the candidates are generated dense in the domain (recall the earlier mentioned [56, Theorem 1.3]).

## IV. EXPERIMENTS

We compare the efficiency and accuracy of our solver to the efficiency and accuracy of several optimization methods in synthetic (Sec. IV-B) and realistic experiments (Sec. IV-C). Additionally, the realistic camera placement examples (Sec. IV-C) show the actual computing time of maximizing an expensive objective function by the proposed solver. Let us start by consulting on the implementation of the Algorithms (Sec. IV-A).

### A. Implementation:

In our algorithm, a single global surrogate $\bar{f}$ is used to cache all the evaluation points. Experiments (Fig. 6) showed that a global search in Line 4 and update to a global surrogate in Lines 5 and 13 are necessary for the convergence of the surrogate. Using a surrogate on each subspace with the same dimension as the global surrogate are difficult: Rarely, linear dependent evaluation points from the subspaces yield a singular matrix (6). More probable is the independent case, in which all the evaluation points in a subspace still shape a hyperline in the domain. The surrogate shapes a smooth hill on the hyperline even though the expensive objective $f$ does not have an extremum there.

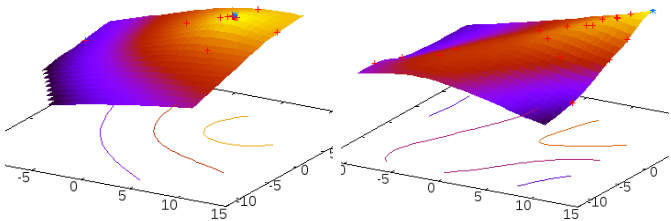

Fig. 6. Surrogate upon termination of the BCA that optimized the function of Fig. 5 (left) without update Lines 4 and 5 in Algorithm 3 and (right) with update line.

In order to find the furthest distance in Line 1 of the Search-Algorithm 2 an MMA implemented by the nlopt library is used. To pick an initial differentiable point for Line 1, a point is randomly chosen in the domain excluding the previous variable vectors. The constraints of the search (Line 3) need to be modified before each optimization, which is done with the IPOPT library. In Line 2 of Alg.1, the equation system is solved by an LU-decompos. of the gsl library.

In the synthetic examples, the solver parameters are set such, that none of the solver iterations exceeded the maximum number of function evaluations, while in the realistic examples the solvers were purposely interrupted to record the time consumption. As termination criterion of Algorithm 3 in Line 2 we use $\Delta_0 = 5$ in case of the synthetic examples and $\Delta_0 = 0.7$ in case of the realistic examples. The number of evaluations of the actual objective function (in the Update-Algorithm 1 in Line 1) are set to 2000 in the synthetic examples, and 50 in the realistic examples. The size $n \in \mathbb{N}$ of the problem was 2,3,4, and 5 in the synthetic examples and 24 in the realistic examples. The cycle pattern $<\beta_1, \ldots, \beta_L> = <0.98, 0.6, 0.75, 0.2, 0.01>$ is used in Line 3 of Algorithm 3.

### B. Synthetic Examples:

In order to solve the optimal camera placement problem, [41] has used an evolutionary method, [8] has used hill climbing, [39], [40] have used greedy methods. We compare the results of our method to the results of solvers similar to these implemented by the nlopt library. For local hill climbing some do not use derivative information (LN_NELDERMEAD), some more sophisticated do (LD_MMA and LD_SLSQP). As a greedy method, a local method that solves subspace maximizations (LN_SBPLX) is used and as a global evolutionary method the solver GN_ISRES. We compare both quantized (Fig. 4, left) and piecewise differentiable functions (Fig. 7, (3)) in several dimensions $n = 2, \ldots, 5$.

The proposed methods have randomized components. The output could therefore differ even if the same solver parameters are used. This is the reason why we called each optimization method in groups of 20 solver calls with the same solver configuration but different initial variable vectors randomly chosen from the interval $[4, 9)$ not containing the optimum at $(-2, \ldots, -2)$. The investigated criteria are then minimized, averaged, and maximized over the test runs in one group. [Algorithm 3 needs $(n+1)$ variable vectors which are chosen in a simplex of the domain starting at the initial vector].

Please note the *logarithmic scale* in the following plots which is why the results of the other local solvers are even less accurate in reality, the number of function evaluations of the other global solver is even higher and values near 0 (GN_ISRES) are not shown. The deviation of the result of the local solvers from the actual maximum value exceeds the final deviation of the RBF-BCA result by about 30 on average (Fig. 7 (1)). The number of function evaluations of the global solver exceeds the number of evaluations during the RBF-BCA about 30 times on average. (Fig. 7 (2)).

The RBF-BCA was also compared to the RBF-solver in [77] combining the RBF and EAM without BCA (Fig. 7 (4)-(5)):

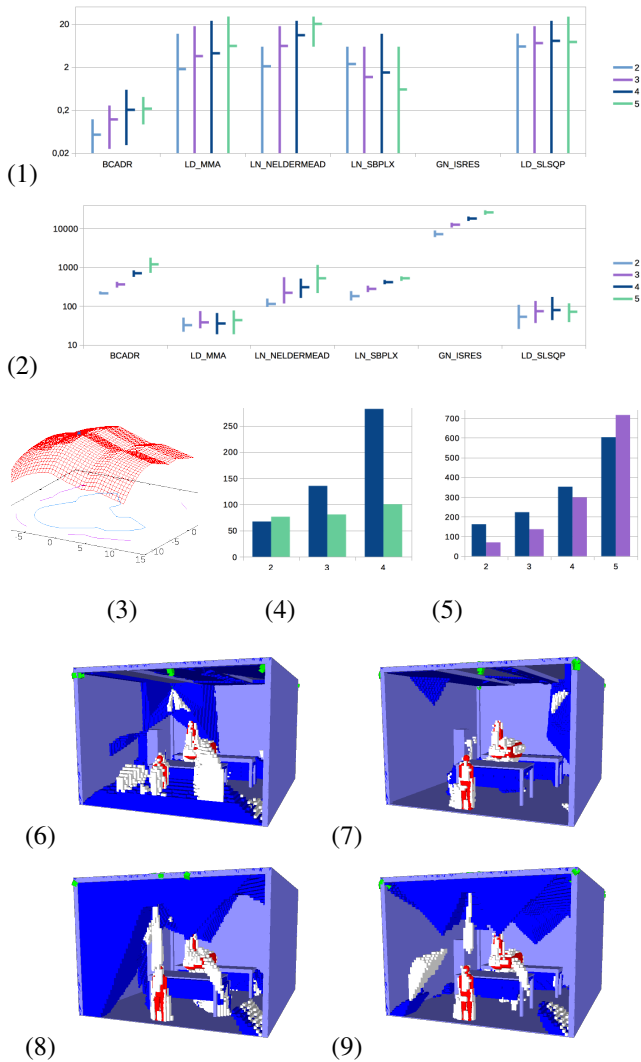

Fig. 7. Comparison of solvers for optimal camera placement: Bar plot of the deviation between final result of a solver and actual maximum (1, smaller = more accurate) and bar plot of the number of function calls (2, smaller = faster) maximizing the function (3) in $n = 2, \ldots, 5$ dimensions: Each bar is drawn from the maximum to the minimum including a mark at the average of each group of 20 solver calls. Similar results have been found for the quantized function of Fig. 4.
(4-5): Number of function evaluations (ordinate) of the RBF-BCA (green/purple) in $n = 2, \ldots, 4/5$ dimensions with $M = 2, \ldots, 4/5$ subspaces (abscissa) compared to the RBF-Solver without BCA component (blue). (4) Incorporating symmetrical sample pairs (green) (5) Sequential function evaluations (purple) of the parallelized RBF-BCA.
(6)-(8): Final qualitative results of the realistic example in Fig. 1 solved by the RBF-BCA (6, 85.28%, 1.5 min), the RBF-Solver [77] (7, 84.63%, 2 min), the LN_SBPLX (nlopt library) (8, 79.88%, 1.5 min), and placement in the corners of the room (9, 81.72%, consideration time) (The random placement (47%) is not illustrated for condensation.)

The incorporation of symmetric sample pairs in the surrogate increases the efficiency of our EAM linearly. Additionally, the number of *sequential* function evaluations of the RBF-Solver according to the complexity (13) exceeds the number of *sequential* function calls of the parallelized RBF-BCA (3 evaluations of $(n+2)$ were sequential).

*C. Realistic Examples*

Six cameras were placed and oriented (four variables each) with our method, the most promising state of the art solver (LN_SBPLX), the RBF-Solver, a heuristic to place the cameras in the corners of the room, and a random placement.

The quantized, expensive black-box objective function renders images (320×240 pixels, OpenGL version 2.2.0, NVIDIA 340.46), simulates a visual hull algorithm (80×90×75 voxels) in the environment illustrated in Fig. 1 (85k human faces, 35k robot faces, and 0.3k environment faces) and measures the volume of the voxels that are not in the visual hull. The experiments have been performed on an "Intel(R) Core(TM)2 Duo CPU E8400" with 1999.0MHz and 6144 KB cache size using the 64-bit operating system "openSUSE 13.1 (x86_64)" and 2GB RAM.

The proposed solver finds the best solution of all methods and accelerates the objective function (Fig. 7 (6)-(9)) the same way the LN_Sbplx does seeing that it also calls the objective on subspaces of the domain. With our methods we have even found mounting spots with several cameras pointing to different directions to enlarge the opening angle of the cameras, a problem called *angular coverage* in [44]. Most of the cameras in all the environments are placed at the boundary of the domain, namely the walls, floor, or edges of the room. The few remaining camera positions are non-intuitive.

## V. CONCLUSION

The manual placement of multiple cameras in environments with several rooms or complicated obstacles takes consideration time and becomes less intuitive the more complex the environment becomes. We strongly recommend an iterative method, in particular the combination of a BCA to accelerate the objective function, a surrogate for fast, smooth optimization, and a EAM for global convergence.

## REFERENCES

[1] Zhou, D., Schwager, M.: Virtual rigid bodies for coordinated agile maneuvering of teams of micro aerial vehicles. In: Robotics and Automation (ICRA), IEEE (2015) 1737–1742
[2] Cheung, G., Kanade, T., Bouguet, J.Y., Holler, M.: A real time system for robust 3d voxel reconstruction of human motions. In: Computer Vision and Pattern Recognition, 2000. Proceedings. IEEE Conference on. Volume 2., IEEE (2000) 714–720
[3] Li, C., Yu, L., Fei, S.: Large-scale, real-time 3d scene reconstruction using visual and imu sensors. IEEE Sensors Journal **20**(10) (2020) 5597–5605
[4] Werner, T., Henrich, D.: Efficient and precise multi-camera reconstruction. In: Proceedings of the International Conference on Distributed Smart Cameras, ACM (2014) 23
[5] Shahbaz, A., Jo, K.H.: Improved change detector using dual-camera sensors for intelligent surveillance systems. IEEE Sensors Journal (2020)
[6] Shih, H.C.: A survey of content-aware video analysis for sports. IEEE Transactions on Circuits and Systems for Video Technology **28**(5) (2017) 1212–1231
[7] Mittal, A., Davis, L.: A general method for sensor planning in multi-sensor systems: Extension to random occlusion. International Journal of Computer Vision **76**(1) (2008) 31–52
[8] Bodor, R., Drenner, A., Schrater, P., Papanikolopoulos, N.: Optimal camera placement for automated surveillance tasks. Journal of Intelligent and Robotic Systems **50**(3) (2007) 257–295
[9] Hänel, M., Kuhn, S., Henrich, D., Grüne, L., Pannek, J.: Optimal camera placement to measure distances regarding static and dynamic obstacles. International Journal of Sensor Networks **12**(1) (2012) 25–36
[10] O'Rourke, J.: Art gallery theorems and algorithms. Volume 1092. Oxford University Press Oxford (1987)

[11] Erdem, U.M., Sclaroff, S.: Optimal placement of cameras in floorplans to satisfy task requirements and cost constraints. In: OMNIVIS workshop. Volume 4. (2004)

[12] Becker, E., Guerra-Filho, G., Makedon, F.: Automatic sensor placement in a 3D volume. In: Proceedings of the 2nd International Conference on Pervasive Technologies Related to Assistive Environments, ACM (2009) 36

[13] Yabuta, K., Kitazawa, H.: Optimum camera placement considering camera specification for security monitoring. In: IEEE International Symposium on Circuits and Systems, 2008. ISCAS 2008. (2008) 2114–2117

[14] Marzal, J.: The three-dimensional art gallery problem and its solutions. PhD thesis, Murdoch University (2012)

[15] Mittal, A., Davis, L.: Visibility analysis and sensor planning in dynamic environments. In: Computer Vision-ECCV 2004. Springer (2004) 175–189

[16] Mittal, A.: Generalized multi-sensor planning. In: Computer Vision–ECCV 2006. Springer (2006) 522–535

[17] Goodchild, M., Lee, J.: Coverage problems and visibility regions on topographic surfaces. Annals of Operations Research **18**(1) (1989) 175–186

[18] Hörster, E., Lienhart, R.: On the optimal placement of multiple visual sensors. In: Proceedings of the 4th ACM international workshop on Video surveillance and sensor networks, ACM (2006) 111–120

[19] Holt, R.J., Man, H., Martini, R., Mukherjee, I., Netravali, R., Wang, J.: Summary of results on optimal camera placement for boundary monitoring. In: Data Mining, Intrusion Detection, Information Assurance, and Data Networks Security 2007. Volume 6570., International Society for Optics and Photonics (2007) 657005

[20] Fiore, L., Somasundaram, G., Drenner, A., Papanikolopoulos, N.: Optimal camera placement with adaptation to dynamic scenes. In: IEEE International Conference on Robotics and Automation. (2008) 956–961

[21] Church, R., Velle, C.: The maximal covering location problem. Papers in regional science **32**(1) (1974) 101–118

[22] Fiore, L., Fehr, D., Bodor, R., Drenner, A., Somasundaram, G., Papanikolopoulos, N.: Multi-camera human activity monitoring. J Intell Robot Syst **52**(1) (2008) 5–43

[23] Murray, A., Kim, K., Davis, J., Machiraju, R., Parent, R.: Coverage optimization to support security monitoring. Computers, Environment and Urban Systems **31**(2) (2007) 133–147

[24] Motai, Y., Kosaka, A.: Hand–eye calibration applied to viewpoint selection for robotic vision. Industrial Electronics, IEEE Transactions on **55**(10) (2008) 3731–3741

[25] Baumann, M., Léonard, S., Croft, E., Little, J.: Path planning for improved visibility using a probabilistic road map. Robotics, IEEE Transactions on **26**(1) (2010) 195–200

[26] Abrams, S., Allen, P., Tarabanis, K.: Computing camera viewpoints in an active robot work cell. International Journal of Robotics Research **18**(3) (1999) 267–285

[27] Reed, M., Allen, P.: Constraint-based sensor planning for scene modeling. Pattern Analysis and Machine Intelligence, IEEE Transactions on **22**(12) (2000) 1460–1467

[28] Banta, J., Wong, L., Dumont, C., Abidi, M.: A next-best-view system for autonomous 3-D object reconstruction. Systems, Man and Cybernetics, Part A: Systems and Humans, IEEE Transactions on **30**(5) (2000) 589–598

[29] Barreto, J., Perdigoto, L., Caseiro, R., Araujo, H.: Active stereo tracking of targets using line scan cameras. Robotics, IEEE Transactions on **26**(3) (2010) 442–457

[30] Song, B., Soto, C., Roy-Chowdhury, A., Farrell, J.: Decentralized camera network control using game theory. In: Second ACM/IEEE International Conference on Distributed Smart Cameras. (2008)

[31] Marchand, É., Spindler, F., Chaumette, F.: ViSP for visual servoing: A generic software platform with a wide class of robot control skills. Robotics & Automation Magazine, IEEE **12**(4) (2005) 40–52

[32] Kaminka, G., Schechter-Glick, R., Sadov, V.: Using sensor morphology for multirobot formations. Robotics, IEEE Transactions on **24**(2) (2008) 271–282

[33] Ercan, A., Yang, D., Gamal, A., Guibas, L.: Optimal placement and selection of camera network nodes for target localization. In: Lecture notes in computer science. Springer Verlag Berlin Heidelberg (2006) 389–404

[34] Yang, D., Shin, J., Guibas, L., Ercan, A.: Sensor tasking for occupancy reasoning in a network of cameras. In: Proceedings of 2nd IEEE International Conference on Broadband Communications, Networks and Systems (BaseNets' 04), Citeseer (2004)

[35] Shih, S.E., Tsai, W.H.: Optimal design and placement of omni-cameras in binocular vision systems for accurate 3-d data measurement. IEEE transactions on circuits and systems for video technology **23**(11) (2013) 1911–1926

[36] Ercan, A., Gamal, A., Guibas, L.: Camera network node selection for target localization in the presence of occlusions. In: In SenSys Workshop on Distributed Cameras. (2006)

[37] Ercan, A.O.: Object tracking via a collaborative camera network. PhD thesis, Stanford University (2007)

[38] Altahir, A.A., Asirvadam, V.S., Hamid, N.H.B., Sebastian, P., Hassan, M.A., Saad, N.B., Ibrahim, R., Dass, S.C.: Visual sensor placement based on risk maps. IEEE Transactions on Instrumentation and Measurement **69**(6) (2019) 3109–3117

[39] Altahir, A.A., Asirvadam, V.S., Hamid, N.H., Sebastian, P., Saad, N., Ibrahim, R., Dass, S.C.: Modeling multicamera coverage for placement optimization. IEEE sensors letters **1**(6) (2017) 1–4

[40] Suresh, S., Narayanan, A., Menon, V.: Maximizing camera coverage in multi-camera surveillance networks. IEEE Sensors Journal (2020)

[41] Morsly, Y., Aouf, N., Djouadi, M.S., Richardson, M.: Particle swarm optimization inspired probability algorithm for optimal camera network placement. IEEE Sensors Journal **12**(5) (2011) 1402–1412

[42] Zhao, J., Yoshida, R., Cheung, S.c.S., Haws, D.: Approximate techniques in solving optimal camera placement problems. International Journal of Distributed Sensor Networks **9**(11) (2013) 241913

[43] Gonzalez-Barbosa, J.J., García-Ramírez, T., Salas, J., Hurtado-Ramos, J.B., et al.: Optimal camera placement for total coverage. In: 2009 IEEE International Conference on Robotics and Automation, IEEE (2009) 844–848

[44] Yildiz, E., Akkaya, K., Sisikoglu, E., Sir, M.Y.: Optimal camera placement for providing angular coverage in wireless video sensor networks. IEEE transactions on computers **63**(7) (2013) 1812–1825

[45] Brévilliers, M., Lepagnot, J., Idoumghar, L., Rebai, M., Kritter, J.: Hybrid differential evolution algorithms for the optimal camera placement problem. Journal of Systems and Information Technology (2018)

[46] Zhao, J., Sen-ching, S.C.: Multi-camera surveillance with visual tagging and generic camera placement. (2007) 259–266

[47] Polyak, B., Fatkhullin, I.: Use of projective coordinate descent in the fekete problem. Computational Mathematics and Mathematical Physics **60**(5) (2020) 795–807

[48] Aardal, K., van den Berg, P., Gijswijt, D., Li, S.: Approximation algorithms for hard capacitated k-facility location problems. European Journal of Operational Research **242**(2) (2015) 358–368

[49] Abidi, B., Aragam, N., Yao, Y., Abidi, M.: Survey and analysis of multimodal sensor planning and integration for wide area surveillance. ACM Computing Surveys (CSUR) **41**(1) (2008) 1–36

[50] Nelder, J., Mead, R.: A simplex method for function minimization. The computer journal **7**(4) (1965) 308–313

[51] Box, M.: A new method of constrained optimization and a comparison with other methods. The Computer Journal **8**(1) (1965) 42–52

[52] Wächter, A., Biegler, L.: On the implementation of an interior-point filter line-search algorithm for large-scale nonlinear programming. Mathematical programming **106**(1) (2006) 25–57

[53] Svanberg, K.: A class of globally convergent optimization methods based on conservative convex separable approximations. SIAM journal on optimization **12**(2) (2002) 555–573

[54] Kraft, D.: Algorithm 733: TOMP–Fortran modules for optimal control calculations. ACM Transactions on Mathematical Software (TOMS) **20**(3) (1994) 262–281

[55] Schittkowski, K.: A robust implementation of a sequential quadratic programming algorithm with successive error restoration. Optimization Letters **5**(2) (2011) 283–296

[56] Törn, A., Zilinskas, A.: Global optimization, lecture notes in computer science. Volume 350. Springer-Verlag, Berlin (1989)

[57] Schlüter, M., Egea, J., Antelo, L., Alonso, A., Banga, J.: An extended ant colony optimizaion algorithm for integrated process and control system design. Ind. Eng. Chem. **48**(14) (2009) 6723–6738

[58] Bertsimas, D., Tsitsiklis, J., et al.: Simulated annealing. Statistical science **8**(1) (1993) 10–15

[59] Fornasier, M., Schönlieb, C.B.: Subspace correction methods for total variation and $\ell_1$-minimization. SIAM Journal on Numerical Analysis **47**(5) (2009) 3397–3428

[60] Qu, Z., Richtárik, P., Takác, M., Fercoq, O.: Sdna: Stochastic dual newton ascent for empirical risk minimization. In: International Conference on Machine Learning. (2016) 1823–1832

[61] Fornasier, M., Langer, A., Schönlieb, C.B.: A convergent overlapping domain decomposition method for total variation minimization. Numerische Mathematik **116**(4) (2010) 645–685

[62] Bertsekas, D., Tsitsiklis, J.: Parallel and distributed computation: Numerical methods. Athena Scientific (1997)

[63] Mohlenkamp, M., Young, T.R., Bárány, B.: Transient dynamics of block coordinate descent in a valley. International Journal of Numerical Analysis & Modeling **17**(4) (2020)

[64] Grippo, L., Sciandrone, M.: Globally convergent block-coordinate techniques for unconstrained optimization. Optimization methods and software **10**(4) (1999) 587–637

[65] Nesterov, Y.: Efficiency of coordinate descent methods on huge-scale optimization problems. SIAM Journal on Optimization **22**(2) (2012) 341–362

[66] Richtárik, P., Takáč, M.: Iteration complexity of randomized block-coordinate descent methods for minimizing a composite function. Mathematical Programming **144**(1-2) (2014) 1–38

[67] Buhmann, M.: Radial basis functions: Theory and implementations. Volume 5. Cambridge university press Cambridge (2003)

[68] Schwarz, H., Köckler, N.: Numerische Mathematik. Springer DE (2009)

[69] Edelsbrunner, H.: Algorithms in combinatorial geometry. Volume 10. Springer (1987)

[70] Duchon, J.: Splines minimizing rotation-invariant semi-norms in Sobolev spaces. In: Constructive theory of functions of several variables. Springer (1977) 85–100

[71] Powell, M.: Recent research at Cambridge on radial basis functions. Springer (1999)

[72] Hardy, R.: Multiquadric equations of topography and other irregular surfaces. Journal of geophysical research **76**(8) (1971) 1905–1915

[73] Björkman, M., Holmström, K.: Global optimization of costly nonconvex functions using radial basis functions. Optimization and Engineering **1**(4) (2000) 373–397

[74] McDonald, D., Grantham, W., Tabor, W., Murphy, M.: Global and local optimization using radial basis function response surface models. Applied Mathematical Modelling **31**(10) (2007) 2095–2110

[75] Regis, R.G., Shoemaker, C.A.: Combining radial basis function surrogates and dynamic coordinate search in high-dimensional expensive black-box optimization. Engineering Optimization **45**(5) (2013) 529–555

[76] Müller, J., Krityakierne, T., Shoemaker, C.: SO-MODS: Optimization for high dimensional computationally expensive multi-modal functions with surrogate search. In: Evolutionary Computation (CEC), 2014 IEEE Congress on, IEEE (2014) 1092–1099

[77] Regis, R., Shoemaker, C.: Constrained global optimization of expensive black box functions using radial basis functions. Journal of Global Optimization **31** (2005) 153–171

[78] Hänel, M.: A Matter of Perspective - Three Dimensional Placement of Multiple Sensors for Change Detection Systems. PhD thesis, University of Bayreuth (2015)

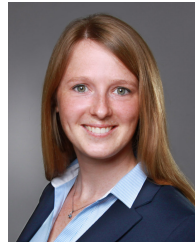

**Dr. Maria L. Hänel** completed her graduate studies in Mathematics with minors in Engineering at the chair for Numerical Mathematics, Optimization, and System and Control Theory at the University of Bayreuth in Germany. She achieved the degree Dr. rer. nat. on the subject of optimal camera placement at the chair for Robotics and Embedded Systems in Bayreuth in 2015.

She received an award for the best grades including Mathematics in high school and achieved a doctoral and post-doctoral scholarship from the DAAD and the University of Bayreuth. The first allowed a five months research stay to colaborate with Prof. Schönlieb on the topic of combining domain decomposition and a surrogate technique at the Cambridge Image Analysis group at the University of Cambridge, UK. In her post-doctoral work, she collaborate with Prof. Knauer of the Algorithms and Data Structures group at the University of Bayreuth on the analysis of vision functions. Afterwards, she conducted research on line scan cameras for the quality control of printing presses in Hamburg, Germany.

She is fascinated by the interplay of complex mathematical theory and practical applications such as optimal camera placement and other problems that include a geometrical component. The latter makes them highly visual and allows to derive and to understand conceptually demanding algorithms, such as interpolation methods, domain decomposition methods and other methods for the optimization of real world problems.

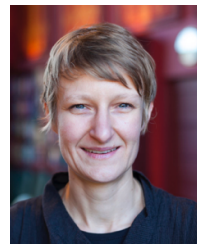

**Prof. Carola-Bibiane Schönlieb** has earned her Master's degree in Mathematics with Distinction at the University of Salzburg in Austria in 2004. She received the degree Doctor of Philosophy at the University of Cambridge, UK, in 2009.

She is currently a Professor of Applied Mathematics and head of the Cambridge Image Analysis (CIA) group at the Department of Applied Mathematics and Theoretical Physics (DAMTP), University of Cambridge, UK. She is the Director of the Cantab Capital Institute for the Mathematics of Information, Director of the EPSRC Centre for Mathematical and Statistical Analysis of Multimodal Clinical Imaging, a Fellow of Jesus College, Cambridge and co-leader of the IMAGES network. Her main research interests are domain decomposition methods, higher-order PDEs, restoration of artwork, image inpainting, high-resolution magnetic resonance imaging and emission tomography, sparse and higher-order variational and PDE regularization, and bilevel optimization for noise estimation.

Prof. Schönlieb has won the Hans-Stegbuchner-Award from the Department of Mathematics at the University of Salzburg, the Mary Bradburn Award from the BFWG, the INiTS Award from INiTS (Innovation into Business) in Vienna and the 3rd Prize in the Category General Technologies, the EPSRC Science Photo Award with a 1st Prize in the Category People, the Whitehead prize of the London Mathematical Society, and the Philip Leverhulme Prize. She received a scholarship from the University of Salzburg (Austria) for exceptional achievements as a student.